\documentclass{Interspeech}

\interspeechcameraready

\title{Beyond Hard Sharing: Efficient Multi-Task Speech-to-Text Modeling with Supervised Mixture of Experts}

\author[affiliation={}, equalcontribution]{Hojun}{Jin}
\author[affiliation={}, equalcontribution]{Eunsoo}{Hong}
\author[affiliation={}]{Ziwon}{Hyung}
\author[affiliation={}]{Sungjun}{Lim}
\author[affiliation={}]{Seungjin}{Lee}
\author[affiliation={}]{Keunseok}{Cho}

\affiliation[nocounter]{}{Samsung Research}{Korea}
\email{hojun.jin@samsung.com, eunsoo1.hong@samsung.com, ziwon.hyung@samsung.com, sungjunj.lim@samsung.com, sjsr.lee@samsung.com, ks1.cho@samsung.com}

\usepackage{comment}
\usepackage{multirow}
\usepackage{float}
\usepackage{kotex}

\begin{document}

\maketitle

\begin{abstract}

Hard-parameter sharing is a common strategy to train a single model jointly across diverse tasks. However, this often leads to task interference, impeding overall model performance. To address the issue, we propose a simple yet effective Supervised Mixture of Experts (S-MoE). Unlike traditional Mixture of Experts models, S-MoE eliminates the need for training gating functions by utilizing special guiding tokens to route each task to its designated expert. By assigning each task to a separate feedforward network, S-MoE overcomes the limitations of hard-parameter sharing. We further apply S-MoE to a speech-to-text model, enabling the model to process mixed-bandwidth input while jointly performing automatic speech recognition (ASR) and speech translation (ST). Experimental results demonstrate the effectiveness of the proposed S-MoE, achieving a 6.35\% relative improvement in Word Error Rate (WER) when applied to both the encoder and decoder.

\end{abstract}

\begin{figure}[t]
  \centering
  \includegraphics[width=\linewidth]{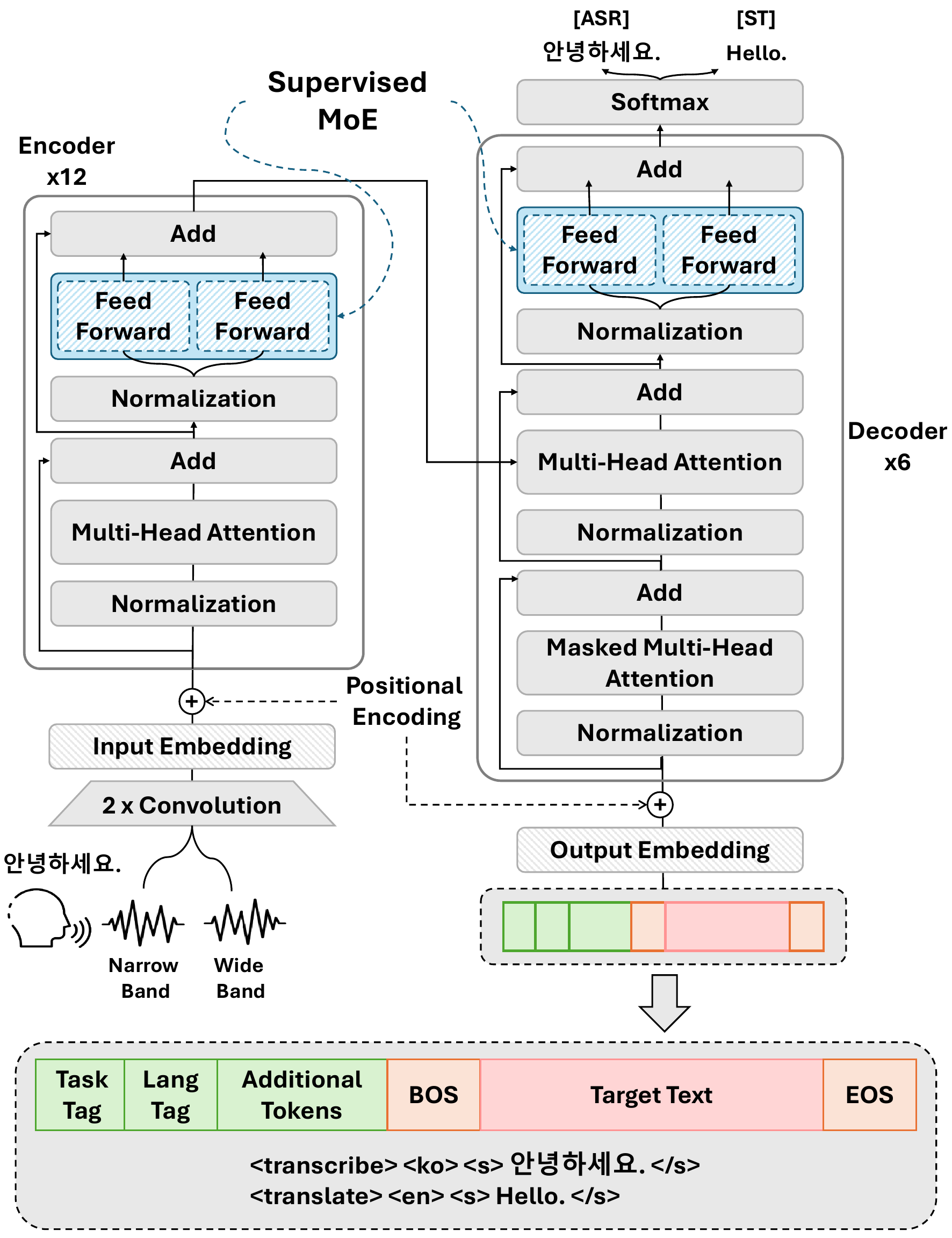}
  \caption{The S-MoE layer embedded within a STT model. The S-MoE in the encoder directs inputs to expert networks based on the speech bandwidth (NB or WB), while the S-MoE in the decoder routes inputs according to the task (ASR or ST).}
  \label{fig:fig1}
\end{figure}

\section{Introduction}

Speech-to-text (STT) models are typically trained on wideband (WB) audio, as it provides richer acoustic features. However, narrowband (NB) audio remains essential in mobile environments, particularly for telephony applications. Due to the spectral differences resulting from the sampling rate gap, NB and WB models are often trained separately. Additionally, many speech-to-text applications require performing both automatic speech recognition (ASR) and speech translation (ST). From a deployment perspective, maintaining multiple models for each task is impractical due to resource constraints on mobile and embedded devices.

Multi-task learning (MTL) is a promising solution to this challenge, enabling a single model to learn shared representations across varying inputs and tasks \cite{crawshaw2020multi, ruder2017overview, zhang2021survey}. Conventional MTL methods that use hard-parameter sharing often suffer from task interference, where the optimization of one task degrades the performance of others \cite{crawshaw2020multi}. 

Many methods have been proposed to mitigate such interference and one effective approach is using modular architectures like Mixture of Experts (MoE) \cite{gupta2022sparsely}. MoE has gained attention due to its ability to allocate specialized parameters for different tasks. However, traditional MoE models rely heavily on gating functions that dynamically route inputs to specific experts. 

In this paper, we propose a novel approach called Supervised Mixture of Experts (S-MoE), specifically designed for efficient multi-task STT training. Our approach is inspired by the MoE principle but tailored for scenarios where both input and output conditions are clearly known. We focus on structured, multi-condition tasks where supervised routing offers a simple and effective solution. Unlike standard MoE models, our method eliminates the need for training gating functions by introducing special guiding tokens that explicitly route tasks to dedicated expert networks. This approach simplifies the routing mechanism while ensuring the model operates with no additional computational overhead, maintaining efficiency during both training and inference.

The contributions of this paper are as follows:
\begin{itemize}
\item We propose an S-MoE approach that mitigates task interference in MTL, using guiding tokens for task routing which minimizes computational cost.
\item We further apply the S-MoE structure to a STT model, enabling it to process mixed-bandwidth input while jointly performing ASR and ST.
\item Our method achieves a 6.35\% improvement in WER when applied to both the encoder and decoder.
\end{itemize}
\section{Approach}

In this section, we propose an architecture as shown in Figure~\ref{fig:fig1}. Further details are presented in the following subsections.

\subsection{Supervised Mixture of Experts (S-MoE)}

By slightly adapting the conventional MoE \cite{shazeer2017outrageously} structure, the output $y$ of the S-MoE module is expressed as Equation \ref{equation:eq1}. Similar to the standard MoE layer, $E_i(x)$ represents the output of $i$-th expert network given an input $x$. The key distinction of S-MoE lies in the gating mechanism. Unlike traditional MoE models that rely on a learnable gating network $G$, our approach employs a predefined gating function $G'$, eliminating the need for additional gating network training. As shown in Figure \ref{fig:fig1}, the number of expert networks $n$ is fixed to $2$ in both the encoder and decoder throughout this work.

\begin{align}
    y &= \sum_{i=0}^{n-1}G'(x)_{i}E_{i}(x)
    \label{equation:eq1}
\end{align}

\subsubsection{Gating function for Encoder S-MoE}
\label{gating-enc}

Models trained on NB tend to underperform when processing WB speech \cite{moreno1994sources,seltzer2006training}. To effectively handle both NB and WB signals within a single model, we incorporate an S-MoE architecture in the encoder. In this design, the feedforward network (FFN) serves as a specialized expert for each bandwidth, while the remaining components in the encoder block are shared across NB and WB inputs. Bandwidth information is pre-labeled for each input, allowing the gating function $G'(x)$ to selectively activate the appropriate expert. Similar to the standard MoE, if $G'(x)_i = 0$, the corresponding expert $E_i(x)$ is not computed. Consequently, WB signals are processed by expert $E_0$, whereas NB signals are handled by expert $E_1$. The gating function of the encoder S-MoE is formally defined in Equation \ref{equation:eq2}.

\begin{align}
    G'_{enc}(x)_i &=
    \begin{cases}
        G'_{enc}(x)_0 &=
        \begin{cases}
            0, & \text{if $x$ is NB signals} \\
            1, & \text{if $x$ is WB signals} 
        \end{cases} \\
        G'_{enc}(x)_1 &=
        \begin{cases}
            0, & \text{if $x$ is WB signals} \\
            1, & \text{if $x$ is NB signals}            
        \end{cases}
    \end{cases}
    \label{equation:eq2}
\end{align}

\subsubsection{Gating function for Decoder S-MoE}

Similar to the encoder, we apply the S-MoE architecture to the decoder to effectively handle both ASR and ST tasks. To determine the appropriate expert for decoding, task tags are prepended to the text input. Based on the target task, the gating function directs ASR inputs to expert $E_1$ and ST inputs to expert $E_0$. The gating function of the decoder S-MoE is given in Equation \ref{equation:eq3}.

\begin{align}
    G'_{dec}(x)_i &=
    \begin{cases}
        G'_{dec}(x)_0 &=
        \begin{cases}
            0, & \text{if task of $x$ is ASR} \\
            1, & \text{if task of $x$ is ST} 
        \end{cases} \\
        G'_{dec}(x)_1 &=
        \begin{cases}
            0, & \text{if task of $x$ is ST} \\
            1, & \text{if task of $x$ is ASR}            
        \end{cases}
    \end{cases}
    \label{equation:eq3}
\end{align}

\subsection{Embedding Flow of the S-MoE model}
\begin{figure*}[t]
  \centering
  \includegraphics[width=\linewidth]{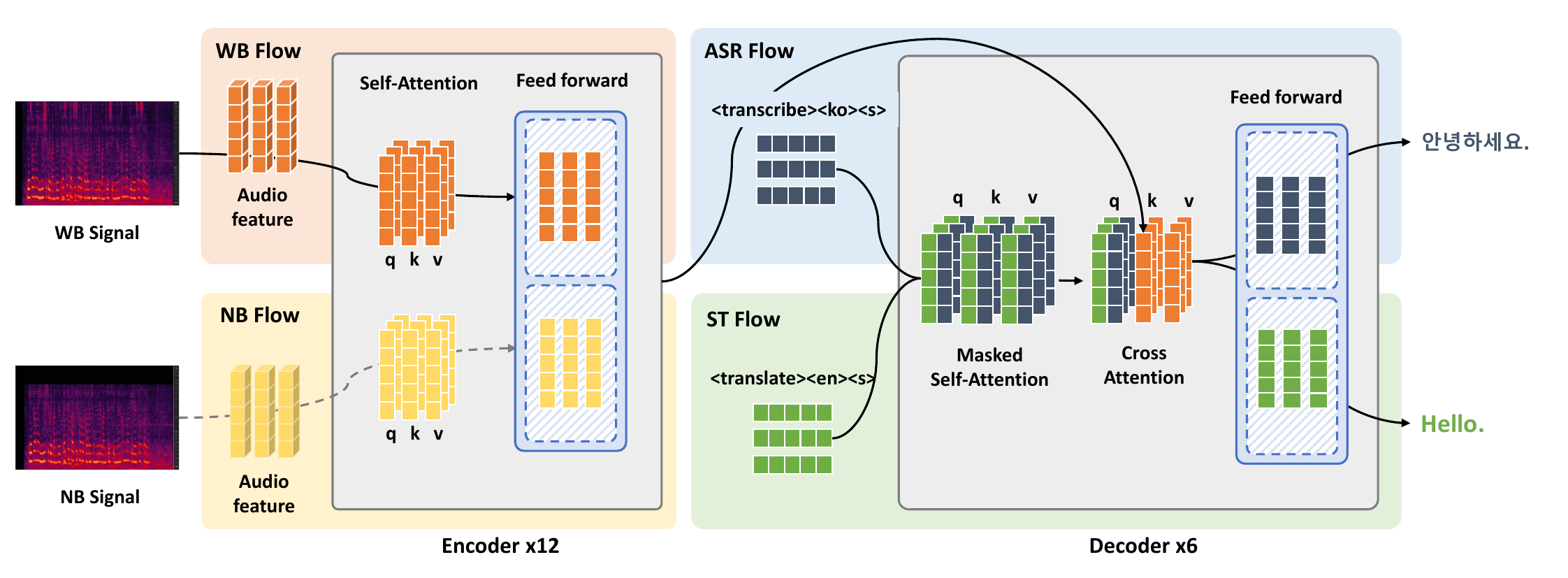}
  \caption{Embedding flow of S-MoE Transformer. During training/inference, a single feedforward block within the encoder is chosen, whereas both feedforward blocks within the decoder are utilized.}
  \label{fig:fig2}
\end{figure*}

As shown in Figure \ref{fig:fig1}, the training process incorporates special tokens to guide the model. At the beginning of each sequence, a task tag (\texttt{<transcribe>} or \texttt{<translate>}) is inserted, followed by a target language tag (\texttt{<en>} or \texttt{<ko>}). Additional special tokens may also be included but not used in this study. The sequence then begins with a \texttt{<beginning\_of\_sentence>} token, followed by the target text. Figure \ref{fig:fig2} illustrates the embedding flow of the proposed S-MoE based multi-task model.

\subsubsection{Training Phase}

The input audio, either NB (\SI{8}{\kilo\hertz}) or WB (\SI{16}{\kilo\hertz}), is first processed and passed into the Transformer encoder. Within the encoder, a FFN block is selected based on the gating function of Encoder S-MoE. The gating function of Encoder S-MoE determines the expert according to the bandwidth of the input signal during training/inference, as in Section \ref{gating-enc}. This means that different FFN blocks are utilized for NB and WB inputs, allowing the model to capture bandwidth-specific representations effectively. In the decoder, two separate FFN blocks are utilized by Decoder S-MoE to process the shared encoded representation. Each training batch consists of samples from a single task, ensuring task-specific optimization. The model is trained with interleaved batches, meaning that both ASR and ST tasks are learned alternately.

\subsubsection{Inference Phase}

During inference, the model can generate ASR and ST outputs simultaneously. By setting the batch size to 2 and assigning the ASR and ST task tags accordingly, both transcription and translation results can be obtained in a single inference step.

\section{Experimental Setup}

\subsection{Datasets}

\subsubsection{Training datasets}
\label{training data}

Our training dataset consists of 40,000 hours of Korean speech data from the publicly available AIHub corpus \cite{aihub_71524, aihub_71260, aihub_123, aihub_463, aihub_71627, aihub_132, aihub_71384, aihub_71379, aihub_464, aihub_568, aihub_71557}. For ASR, we use the provided transcripts as target text. For ST, if a corresponding English translation is not available, we generate the target text using an in-house machine translation (MT) model. If an original translation is provided, we use it directly. As a result, each audio sample has paired ASR and ST data, enabling multi-task learning.

To train the model under different bandwidth conditions, we generate NB and WB variants of the data. WB data is at \SI{16}{\kilo\hertz}, while NB data is at \SI{8}{\kilo\hertz}. We apply two conversion methods: (1) simple downsampling from \SI{16}{\kilo\hertz} to \SI{8}{\kilo\hertz}, and (2) codec-based processing using AMR-WB \cite{bessette2002adaptive}, AMR-NB, and G.711-NB \cite{recommendation1988pulse} codecs. These conversions are applied to 15\% of the training data, creating a NB/WB fine-tuning subset.

\subsubsection{Evaluation datasets}

To evaluate the general performance of ASR and ST, we use two in-house test sets along with the publicly available Fleurs \cite{conneau2023fleurs} dataset. Our in-house test sets consist of 1,000 samples of male and female speech from daily conversations, with reference translations curated by professional translators. Additionally, to assess the model under different bandwidth conditions, we generate NB/WB variants of the test sets using the same conversion methods as in training.

\subsection{Model}
\label{model}

Our base model follows a Transformer-based sequence-to-sequence architecture \cite{vaswani2017attention}, consisting of a 12-layer encoder and a 6-layer decoder.  
We employ sinusoidal positional embeddings and a pre-normalization Transformer structure with GLU activations and 8 attention heads. The FFN inner dimension is set to 2048 and the embedding dimension is 512 with a dropout rate of 0.15. For tokenization, we use a byte-level byte-pair encoding (BBPE) tokenizer with a vocabulary size of 40,000. The input speech is resampled to \SI{16}{\kilo\hertz} and converted into 80-dimensional log Mel filterbank (fbank) features, extracted using a \SI{25}{\milli\second} Hanning window with a \SI{10}{\milli\second} stride. We apply a maximum speech sequence length of 3,000 and a maximum text sequence length of 120, filtering out utterances longer than 30 seconds. The model is trained for 10 epochs with an effective batch size of 3,200, using a cosine annealing scheduler for learning rate decay. The Encoder-Decoder S-MoE model in Section \ref{EncDecS-MoE result} is initialized using the Decoder S-MoE model in Section \ref{DecSMoE result} and trained for 5 epochs with a NB/WB fine-tuning subset.

\subsection{Metric}

For ST, we use the BLEU metric \cite{papineni2002bleu} to measure translation quality. We use the \texttt{sacrebleu} library. For ASR, we report WER, which is a widely used metric in ASR research.

\renewcommand{\arraystretch}{1.1}
\begin{table*}[h!]
    \caption{Impact of Decoder S-MoE on ASR/ST}
    \label{tab:dec-smoe}
    \centering
    \begin{tabular}{|c|c|c|c|c|c|c|c|c|}
    \hline
    \multirow{2}{*}{\textbf{Model}} & \multicolumn{2}{c|}{\textbf{Param}} & \multicolumn{2}{c|}{\textbf{Male TC}} & \multicolumn{2}{c|}{\textbf{Female TC}} & \multicolumn{2}{c|}{\textbf{Fleurs}} \\
    \cline{2-9}
    & Trainable & Active & WER & BLEU & WER & BLEU & WER & BLEU \\
    \hline
    Whisper-base & 74M & 74M & 8.73 & 13.6 & 13.40 & 11.9 & 18.03 & 3.3 \\
    Whisper-small & 244M & 244M & 5.15 & 26.5 & 7.40 & 25.8 & 10.27 & 12.6 \\
    Whisper-large-v3 & 1,550M & 1,550M & 4.46 & 33.7 & 6.87 & 32.8 & 5.32 & 20.1 \\
    \hline
    Ours (Base-Model) & 107M & 107M & 3.93 & 34.3 & 5.36 & 33.2 & 5.46 & 21.1 \\
    Ours (DecFFNx2) & 119M & 119M & 3.83 & 34.1 & 5.36 & 33.5 & 5.46 & 21.5 \\
    \textbf{Ours (DecS-MoE)} & 119M & \textbf{107M} & \textbf{3.53} & \textbf{34.8} & \textbf{4.89} & \textbf{33.9} & \textbf{5.23} & \textbf{21.5} \\
    \hline
    \end{tabular}
\end{table*}

\renewcommand{\arraystretch}{1.1}
\begin{table*}[h!]
    \caption{Impact of Encoder-Decoder S-MoE on ASR/ST}
    \label{tab:enc-dec-smoe}
    \centering
    \begin{tabular}{|c|l|c|c|c|c|c|c|}
    \hline
    \multirow{2}{*}{\textbf{Model}} & \multirow{2}{*}{\textbf{Codec}} & \multicolumn{2}{c|}{\textbf{Male TC}} & \multicolumn{2}{c|}{\textbf{Female TC}} & \multicolumn{2}{c|}{\textbf{Fleurs}} \\
    \cline{3-8}
    & & WER & BLEU & WER & BLEU & WER & BLEU \\
    \hline
     \multirow{5}{*}{Base-Model} & NB & 4.11 & 33.8 & 6.60 & 32.5 & 5.91 & 19.9 \\
      & G.711 NB & 4.18 & 33.9 & 6.77 & 32.4 & 6.50 & 19.9 \\
      & AMR NB & 4.78 & 32.8 & 11.30 & 29.8 & 13.69 & 17.9 \\
      \cline{2-8}
      & WB & 4.51 & 33.8 & \textbf{6.16} & 32.0 & 8.71 & \textbf{19.1} \\
      & AMR WB & 3.98 & 33.9 & 5.46 & 32.9 & 5.84 & 19.8 \\
    \hline
    \multirow{5}{*}{\textbf{EncDecS-MoE}} & NB & \textbf{4.08} & \textbf{34.1} & \textbf{6.22} & \textbf{32.9} & \textbf{5.66} & \textbf{20.4} \\
    & G.711 NB & \textbf{3.92} & \textbf{34.2} & \textbf{5.96} & \textbf{32.8} & \textbf{6.15} & \textbf{20.0} \\  
    & AMR NB & \textbf{4.65} & \textbf{34.0} & \textbf{10.43} & \textbf{30.5} & \textbf{12.96} & \textbf{18.2} \\
    \cline{2-8}
    & WB & \textbf{4.17} & \textbf{34.2} & 6.98 & \textbf{32.3} & \textbf{8.30} & \textbf{19.1} \\    
    & AMR WB & \textbf{3.88} & \textbf{34.2} & \textbf{5.18} & \textbf{33.5} & \textbf{5.34} & \textbf{20.2} \\    
    \hline
    \end{tabular}
\end{table*}

\section{Experimental Results}

\subsection{Decoder S-MoE}
\label{DecSMoE result}
To validate the effectiveness of S-MoE applied to the decoder blocks, we conducted experiments on two tasks: Korean-to-English (ko2en) ST and Korean ASR (ko-ASR). The results are summarized in Table \ref{tab:dec-smoe}. The model described in Section \ref{model} serves as the primary baseline (Base-Model) for comparison.

The trainable parameters of S-MoE increase linearly with the number of experts. According to \cite{shazeer2017outrageously}, despite a slight increase in parameters, the MoE structure retains a computational advantage by activating only a subset of expert networks based on the routing mechanism. In our experiments, we set the number of experts to 2 to support multi-task learning of ASR and ST. The model with S-MoE applied to the decoder is referred to as DecS-MoE in the tables. To match the parameter increase in DecS-MoE, we also trained a variant where the decoder FFN dimension is doubled (DecFFNx2).

For broader comparison, we included OpenAI's Whisper \cite{radford2023robust} models in our experiments. Whisper is a widely used multilingual speech model trained with diverse language pairs. Our models are optimized for a single translation direction so a direct comparison may not be entirely fair, but Whisper provides a strong reference point for assessing ko-ASR and ko2en ST performance.

As shown in Table \ref{tab:dec-smoe}, DecS-MoE consistently outperforms the Base Model across all test sets for both ST and ASR. We observe a 0.5-point increase in BLEU and a 0.4\% decrease in WER on the Male test set. For the Female test set, BLEU increases by 0.7, while WER decreases by 0.47\%. Similarly on the Fleurs test set, DecS-MoE achieves a 0.4 improvement in BLEU and a 0.2\% reduction in WER. Our proposed DecS-MoE achieves an improvement of 0.37\% in WER and 0.53 in BLEU on average over three test sets. This represents a relative improvement of 8.06\% in WER and 1.77\% in BLEU compared to the Base Model. This improvement suggests that applying S-MoE to the decoder enhances performance without introducing negative transfer between tasks. 

Comparing models with the same number of trainable parameters, DecS-MoE even achieves better performance than DecFFNx2. On the Male test set, BLEU improves by 0.7 points, while WER is reduced by 0.3\%. Similarly, for the Female test set, DecS-MoE achieves a 0.4-point increase in BLEU and a 0.47\% drop in WER. In the case of Fleurs, ST performance remains unchanged, but WER decreases by 0.23\%. The averaged results over all test sets show an improvement of 0.33\% in WER and 0.37 in BLEU. This enhancement corresponds to a relative improvement of 7.33\% in WER and 1.22\% in the BLEU score. These results indicate that simply increasing the FFN dimension is not the most effective approach for multi-task learning. Instead, allocating separate experts for each task proves to be a more efficient strategy in terms of both performance and computational cost, as DecS-MoE maintains the same number of active parameters as the Base Model during inference.

Even with the Whisper models, DecS-MoE achieves superior performance across all ASR/ST test sets. Despite having only one-tenth of the trainable parameters compared to Whisper-large (1,550M), DecS-MoE(119M) achieves an average BLEU improvement of 1.2 while reducing WER by 1\%. This demonstrates a relative enhancement of 21.98\% in WER and 3.99\% in BLEU.

\subsection{Encoder-Decoder S-MoE}
\label{EncDecS-MoE result}

We propose an Encoder S-MoE to handle speech signals with varying bandwidths. Specifically, the encoder consists of two experts, one for NB signals and one for WB signals. In Section \ref{DecSMoE result}, we demonstrated that Decoder S-MoE is effective at jointly training ASR and ST tasks. Therefore, we apply S-MoE to both the encoder and the decoder (EncDecS-MoE). With a total of four experts, the EncDecS-MoE model has 144M trainable parameters, whereas the Base-Model has 107M. However, during inference, both models utilize the same number of active parameters (107M). For example, when processing NB speech inputs for ST, the EncDecS-MoE model routes the encoder to the NB expert and the decoder to the ST expert, activating only the relevant experts for efficient computation.

All models in Table \ref{tab:enc-dec-smoe} are fine-tuned on the NB/WB data described in Section \ref{training data}. Since the use of NB speech signals is more limited compared to WB signals, NB does not require as much training data. To train for both NB and WB environments effectively, we fine-tune the pre-trained baseline model, which was initially trained only on WB data.

As shown in Table \ref{tab:enc-dec-smoe}, the EncDecS-MoE demonstrates consistent performance gain in almost all cases. In the NB environment, the EncDecS-MoE achieves an average BLEU score of 28.57, compared to 28.10 for the Base-Model and an average WER of 6.67\%, which is lower than the Base-Model's 7.09\%. This demonstrates a relative improvement of 6.35\% in WER and a 1.63\% gain in BLEU score. Furthermore, under the WB environment, our proposed model achieves a relative improvement of 2.39\% in WER and 1.15\% in BLEU compared to the average performance. NB signals capture only lower frequencies, which may negatively impact training when handling WB and NB data in a single model. However, the results suggest that implementing separate FFNs within the encoder can solve the problem. While using separate experts slightly increases the model size, the number of active parameters remains the same as in the Base-Model. Thus, applying S-MoE in both the encoder and decoder improves performance without compromising inference speed.

\section{Conclusion}

We propose the S-MoE architecture, which mitigates task interference in multi-task learning for STT applications. By using guiding tokens instead of dynamic gating functions, S-MoE ensures efficient training and inference while improving performance across various tasks. Our approach enables a single model to handle ASR and ST tasks with mixed-bandwidth input, making it ideal for resource-constrained environments. Future work will focus on expanding S-MoE to support more speech tasks and multilingual capabilities, further enhancing its real-world applicability.

\bibliographystyle{IEEEtran}
\bibliography{template}

\end{document}